\documentclass[runningheads,a4paper]{llncs}

\usepackage{t1enc}         
\usepackage[english]{babel}
\usepackage{mathptmx}

\newcommand{\bequ}{\begin{quote}}
\newcommand{\enqu}{\end{quote}}

\institute{Department of Computer Science and Engineering\\
  Chalmers University of Technology and University of Gothenburg}

\usepackage{graphicx}
\usepackage{graphicx}
\usepackage{paralist} % needed for compact lists
\usepackage[normalem]{ulem} % needed by strike
\usepackage[urlcolor=black,colorlinks=true,citecolor=black]{hyperref}
\usepackage[utf8]{inputenc}  % char encoding

\title{Embedded Controlled Languages}
\author{Aarne Ranta}
\begin{document}
\maketitle
\mbox{}

\begin{abstract}
Inspired by embedded programming languages, an embedded
CNL (controlled natural language) is a proper fragment
of an entire natural language (its host language), but
it has a parser that recognizes the entire host language. This makes it possible to
process out-of-CNL input and give useful feedback to users, instead of just reporting 
syntax errors. This extended abstract explains the main concepts of embedded CNL
implementation in GF (Grammatical Framework), with examples from machine translation and some
other ongoing work.

\vspace{1mm}


\vspace{1mm}

\textit{Preprint version of paper presented at CNL 2014, Galway.}

\end{abstract}

\section{Introduction}

A controlled natural language (CNL) is a strictly defined fragment of a natural language
\cite{kuhn-2014}. 
As fragments of natural languages, CNLs are analogous to
\textbf{embedded domain-specific languages}, which are fragments of general purpose programming languages
\cite{hudak-1996}.
Such languages have been introduced as an
alternative to traditional \textbf{domain-specific languages} (DSL), which have
their own syntax and semantics, and require therefore a specific learning effort.
An embedded DSL is a part
of a general-purpose programming language, the \textbf{host language}, and is therefore readily usable by 
programmers who already know the host language.  
At its simplest, an embedded DSL is little more than a \textbf{library} in 
the host language. Using the library helps programmers to write compact, efficient, and
correct code in the intended domain. But whenever the library does not provide all 
functionalities wanted, the programmer can leave its
straight-jacket and use the host language directly, of course at her own risk.

Embedding a language fragment in the full language presupposes that a grammar of the full language is available.
In the case of natural languages, this is by no means a trivial matter. On the contrary, it is widely
acknowledged that ``all grammars leak'', which means that any formal grammar defining a natural language is
bound to be either incomplete or overgenerating. As a consequence, defining CNLs \textit{formally} as subsets
of natural languages looks problematic.

However, \textit{if} a grammar of the host language exists, then it is useful to 
define the CNL as an embedded language. It enables us to build systems that provide, at the same time,
the rigour of controlled languages and the comfort of graceful degradation. The user of the
system can be guided to stay inside the controlled language, but she will also be understood, at least
to some extent, if she goes outside it.

In this extended abstract, we will outline some recent work on building controlled languages in the
embedded fashion. Our focus will be on multilingual systems, where the CNL yields
high-quality translation and the host language yields browsing quality. But the same structure should be useful
for other applications of CNLs as well, such as query languages.

In Section 2, we will summarize how CNLs are traditionally defined by using GF, Grammatical Framework.
In Section 3, we will show how they can be converted to embedded CNLs. 
In Section 4, we summarize some on-going work and suggest some more applications.

\section{Defining Controlled Languages in GF}

GF \cite{ranta-2011} is a grammar formalism based on a distinction between \textbf{abstract syntax} and
\textbf{concrete syntax}. The abstract syntax is a system of \textbf{trees}. The concrete syntax is a reversible
mapping from trees to \textbf{strings} and \textbf{records}, reminiscent of \textbf{feature structures} in unification-based
grammar formalisms. The separation between abstract and concrete syntax makes GF grammars \textbf{multilingual},
since one and the same abstract syntax can be equipped with many concrete syntaxes. The abstract syntax is
then usable as an \textbf{interlingua}, which enables \textbf{translation} via \textbf{parsing} the source
language string into a tree followed by the \textbf{linearization} of the tree into the target language.

As an example, consider a predicate expressing the age of a person. The abstract syntax rule is

\begin{verbatim}
  fun aged : Person -> Numeral -> Fact
\end{verbatim}

\noindent
defining a \textbf{function} with the name \texttt{aged}, whose \textbf{type} is a function type of two arguments,
of types \texttt{Person} and \texttt{Numeral}, yielding a value of type \texttt{Fact}. A simple 
concrete syntax rule for English is 

\begin{verbatim}
  lin aged p n = p ++ "is" ++ n ++ "years old"
\end{verbatim}

\noindent
stating that a function application \texttt{(aged p n)} is \textbf{linearized} to the string where the linearization
of \texttt{p} is concatenated (\texttt{++}) with the string \texttt{"is"}, the linearization of \texttt{n}, and the string
\texttt{"years old"}. The corresponding rule for French is

\begin{verbatim}
  lin aged p n = p ++ "a" ++ n ++ "ans"
\end{verbatim}

\noindent
producing sentences literally equivalent to \textit{p has n years}. Thus the concrete syntax allows the
production of different syntactic structures in different languages, while
the abstract syntax form \texttt{(aged p n)}
remains the same, and stands in a compositional relation to the linearizations.

GF is widely used for defining CNLs;
\cite{ranta-angelov-2010,saludes-xambo-2011,ranta-al-2012,davis-al-2012,kaljurand-kuhn-2013} are some examples.
Much of its power comes from the \textbf{resource grammar libraries} (RGL), which
are general purpose grammars enabling GF grammar writers to delegate the ``low-level'' linguistic 
details to generic library code
\cite{ranta-2009,ranta-2009a}. 
If we dig deeper into the concrete syntax of
the \texttt{aged} predicate, we will find lots of such details to cope with:
  number agreement (\textit{one year} vs. \textit{five years}),
  subject-verb agreement: (\textit{I am}, \textit{you are}, \textit{she is}),
  word order (\textit{you are} in declaratives vs. \textit{are you} in questions),
etc; French generally poses harder problems than English.
The use of feature structures instead of plain strings does make it possible to
express all this compositionally in GF. But these details can make the grammar prohibitively difficult
to write, especially since controlled language designers are not always theoretical linguists but
experts in the various domains of application of CNLs. The RGL addresses this problem by providing
general-purpose functions such as

\begin{verbatim}
  mkCl : NP -> VP -> Cl
\end{verbatim}

\noindent
which builds a clause (\texttt{Cl}) from a noun phrase (\texttt{NP}) and a verb phrase (\texttt{VP}) and takes
care of all details of agreement and word order.
The CNL linearization rules can be written
as combinations of such functions. The English rule, in full generality, comes out as compact as

\begin{verbatim}
  lin aged p n = mkCl p (mkVP (mkAP (mkNP n year_N) old_A))
\end{verbatim}

\noindent
and the French,

\begin{verbatim}
  lin aged p n = mkCl p (mkVP avoir_V2 (mkNP n an_N))
\end{verbatim}

\noindent
If a function \texttt{quest} is added to turn facts into questions,
the linearization is in both languages just a simple RGL function call:

\begin{verbatim}
  lin quest fact = mkQS fact
\end{verbatim}

\noindent
The API (application programmer's interface) is the same for all languages in the RGL (currently 29),
but the low-level details that it hides are language-dependent.

\section{Embedding a Controlled Language in the Host Language}

The standard practice in GF is to define CNLs by using 
the RGL rather than low-level hand-crafted linearization rules.
In addition to saving effort, this practice guarantees that the CNL is a valid fragment
of a natural language. The reason is that the RGL is designed only to allow grammatically correct
constructions. 

But what is missing in a CNL built in this way is the rest of the host 
language---the part that is not covered by the RGL rule combinations actually used in the CNL. 
In general, a random sentence has a probability close to zero to
be recognized by the parser. The standard solution is to guide the user
by a predictive editor \cite{khegai-al-2003,ranta-angelov-2010,kuhn-2012}. 
But there are many situations in which this does not 
work so well: for instance, with speech input, or when processing text in batch mode.
Then it can be more appropriate to include not only the CNL but also the host language in the system. 
The system as a whole will then be similar to an embedded DSL with its
general-purpose host language.

The easiest way to combine a CNL with a host language is to
introduce a new start category \texttt{S} with two productions: one
using the CNL start category as its argument type, the other using the host language start category:

\begin{verbatim}
  fun UseCNL  : S_CNL  -> S
  fun UseHost : S_Host -> S
\end{verbatim}

\noindent
The CNL trees can be given priority by biasing the weights of these functions in probabilistic GF
parsing \cite{angelov-ljunglof-2014}. The CNL parse tree will then appear
as the first alternative, whenever it can be found. Since the system sees where the tree comes from, it can
give feedback to the user, for instance by using colours: green colour for CNL trees, yellow for 
host language trees, and red for unanalysed input, as shown (in greyscale in the printed version) in Figure 1.

Since the RGL is designed to guarantee grammaticality, and since all grammars leak, the RGL
does not cover any language entirely. But if the purpose is wide-coverage parsing, we can relax this 
strictness. An easy way to do this is to extend the grammar with \textbf{chunking}.
A chunk can be built from almost any RGL category: sentences, noun phrases, nouns, adjectives, etc:

\begin{verbatim}
  fun ChunkS  : S_Host -> Chunk
  fun ChunkNP : NP     -> Chunk
  fun ChunkN  : N      -> Chunk
\end{verbatim}

\noindent
The top-level grammar has a production that recognizes lists of chunks (\texttt{[Chunk]}):

\begin{verbatim}
  fun UseChunks : [Chunk] -> S
\end{verbatim}

\noindent
It is relatively easy to make the chunking grammar
\textbf{robust}, in the sense that it returns
some analysis for any combination of words. If the input has 
out-of-dictionary words, they can be dealt with by named entity recognition and
morphological guessing. Weights can be set in such a way that longer chunks
are favoured. For instance, \textit{this old city} should be analyzed as one NP chunk rather than
a determiner chunk followed by an adjective chunk and a noun chunk. 
The user can be given feedback, not only by a red
colour indicating that chunks are used, but also by showing the chunk boundaries.

A further step of integration between CNL and the host language is
obtained if CNL sentences are treated as chunks:

\begin{verbatim}
  fun ChunkCNL : S_CNL -> Chunk
\end{verbatim}

\noindent
In the resulting trees, one can use different colours for different chunks inside one
and the same sentence. 

But since both the CNL and the host language are defined in terms of the same RGL structures,
one can take the integration much further, by \textit{intertwining} the CNL and host language
rules. Consider again the CNL predicate

\begin{verbatim}
  fun aged : Person -> Numeral -> Fact
\end{verbatim}

\noindent
where \texttt{Person} and \texttt{Fact} are CNL categories and \texttt{Numeral} is an RGL category. One can
generalize the use of this predicate and other ones by introducing coercions from 
RGL categories to CNL categories used as arguments, 
and from CNL categories used as values to RGL categories (making sure that no
cycles are created):

\begin{verbatim}
  fun np2person : NP -> Person
  fun fact2cl   : Fact -> Cl 
\end{verbatim}

\noindent
The effect of this is seen when we analyse the English sentence
\begin{center}
\textit{John does not believe that the queen is sixty-five years old}
\end{center}
The resulting tree is

\vspace{2mm}

\texttt{mkS negativePol (mkCl John believe\_VS (fact2Cl}

\vspace{-3mm}

\begin{flushright}
\texttt{(}\textbf{aged} \texttt{(np2person (mkNP the\_Det queen\_N))} \textbf{(mkNumeral "65")})))
\end{flushright}
\noindent
where those parts that belong to the CNL are boldfaced. Thus the predicate \texttt{aged} is from the CNL,
but uses as argument \textit{the queen}, which is not in the CNL. The resulting \texttt{Fact} is used as a 
complement to the verb \textit{believe}, which requires an RGL clause. The resulting French translation is
\begin{center}
\textit{John ne croit pas que la reine} \textbf{ait soixante-cinq ans}
\end{center}
which correctly renders the \texttt{aged} idiom defined by the CNL, 
even though its subject is not in the CNL, and even though
the negated main verb puts it into the subjunctive mood, which might
never occur in the CNL itself.

\begin{figure}

\includegraphics[width=1.0\textwidth]{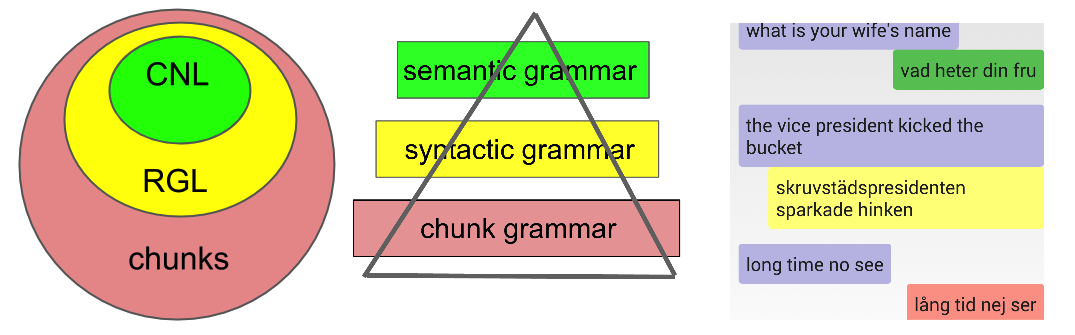}

\caption{
From left: CNL embedded in general purpose RGL embedded in a chunk grammar; 
the corresponding levels in the Vauquois triangle;
a mobile translation application showing the level of confidence in colours (from top:
semantic translation from CNL, syntactic translation from RGL, chunk-based translation).
}
\label{background}
\end{figure}

\section{Work in Progress}

The translation example with the \texttt{aged} predicate
shows that embedded CNL functions introduce semantic structures in translation.
This has been exploited in the wide-coverage GF-based translation system 
\cite{angelov-al-2014}
\footnote{"The Human Language Compiler", \url{http://grammaticalframework.org/demos/app.html}}
where three levels are distinguished by colours: green for the CNL 
(the MOLTO phrasebook \cite{ranta-al-2012}), 
yellow for the RGL syntax,
and red for chunks (Figure 1). These levels correspond to levels in the \textbf{Vauquois triangle}, 
where translation systems are divided on the basis of whether they 
use a semantic interlingua, syntactic transfer, or word-to-word transfer \cite{vauquois-1968}.
The effect of the layered structure with an embedded CNL is that these levels 
coexist in one and the same system. 
The system is currently implemented for 11 languages. Its architecture permits 
easily changing the CNL to some other one and also including several CNLs in a single system. 
At least two experiments are in progress:
one with a mathematical grammar library \cite{saludes-xambo-2011}, another with the multilingual
version of Attempto Controlled English \cite{kaljurand-kuhn-2013}.

In addition to translation, embedded CNLs could be used in query systems, where ``green'' 
parse trees are semantically well-formed queries and other colours are interpreted as key 
phrases and key words. It can also be interesting to match out-of-CNL trees with CNL trees
and try to find the closest semantically complete interpretations. Yet another application
of embedded CNLs would be to implement languages of the ``Simplified English'' type, which 
are not defined by formal grammars but by restrictions posed on the full language \cite{kuhn-2014}.
Such a language could be parsed by using the host language grammar together with a
procedure that checks on the tree level how the restrictions are followed.

\subsection*{Acknowledgement}

Thanks to Krasimir Angelov and Normunds Gruzitis for useful comments.
Swedish Research Council (Vetenskapsrådet) has supported this work under
grant nr. 2012-5746 (Reliable Multilingual Digital Communication).

\bibliographystyle{splncs}
\bibliography{../new-gf-bib}

% LaTeX2e code generated by txt2tags 2.6 (http://txt2tags.org)
% cmdline: txt2tags -ttex embedded.txt
\end{document}